\documentclass[10pt,twocolumn,letterpaper]{article}

\usepackage{cvpr}              

\usepackage[accsupp]{axessibility}  
\usepackage{graphicx}
\usepackage{amsmath}
\usepackage{amssymb}
\usepackage{booktabs}

\usepackage{arydshln} 
\usepackage{multirow}
\usepackage{pifont}
\usepackage{amsmath}
\usepackage{amssymb}
\usepackage{booktabs}
\usepackage{xcolor}
\usepackage{enumitem}
\usepackage{amsfonts}
\usepackage{xparse}
\usepackage{tabu}
\usepackage{subcaption}
\usepackage{wrapfig}

\definecolor{RoseQuartzBg}{HTML}{F7CAC9}
\definecolor{RoseQuartz}{HTML}{F5A798}
\definecolor{Serenity}{HTML}{92A8D1}
\definecolor{OrangeRed}{rgb}{1.0, 0.27, 0.0}
\definecolor{Turquoise}{HTML}{0F4C81}
\definecolor{mint}{rgb}{0.24, 0.71, 0.54}
\definecolor{captioningtask}{HTML}{AE4132}
\definecolor{qatask}{HTML}{0E8088}
\definecolor{temporalprefix}{HTML}{7F00FF}
\definecolor{targettext}{HTML}{3333FF}
\definecolor{prompttext}{HTML}{666666}
\definecolor{videolevel}{HTML}{82B366}
\definecolor{framelevel}{HTML}{6C8EBF}
\definecolor{tokenlevel}{HTML}{D79B00}

\NewDocumentCommand{\heng}
{ mO{} }{\textcolor{OrangeRed}{\textsuperscript{\textit{Heng}}\textsf{\textbf{\small[#1]}}}}

\NewDocumentCommand{\manling}{ mO{} }{\textcolor{blue}{\textsuperscript{\textit{Manling}}\textsf{\textbf{\small[#1]}}}}
\NewDocumentCommand{\xudong}{ mO{} }{\textcolor{blue}{\textsuperscript{\textit{Xudong}}\textsf{\textbf{\small[#1]}}}}

\NewDocumentCommand{\shiyuan}{ mO{} }{\textcolor{purple}{\textsuperscript{\textit{Shiyuan}}\textsf{\textbf{\small[#1]}}}}

\NewDocumentCommand{\s}{ mO{} }{\textcolor{cyan}{\textsuperscript{\textit{Shou}}\textsf{\textbf{\small[#1]}}}}

\usepackage[pagebackref,breaklinks,colorlinks]{hyperref}

\usepackage[capitalize]{cleveref}
\crefname{section}{Sec.}{Secs.}
\Crefname{section}{Section}{Sections}
\Crefname{table}{Table}{Tables}
\crefname{table}{Tab.}{Tabs.}

\def\cvprPaperID{10007} 
\def\confName{CVPR}
\def\confYear{2023}

\begin{document}

\title{Towards Fast Adaptation of Pretrained Contrastive Models for Multi-channel Video-Language Retrieval}

\author{%
  \textbf{
  Xudong Lin\textsuperscript{\textnormal{1}}
Simran Tiwari\textsuperscript{\textnormal{1}}, 
Shiyuan Huang\textsuperscript{\textnormal{1}}, 
 Manling Li\textsuperscript{\textnormal{2}}
}
\\
 \textbf{
Mike Zheng Shou\textsuperscript{\textnormal{3}}, 
Heng Ji\textsuperscript{\textnormal{2}},
Shih-Fu Chang\textsuperscript{\textnormal{1}}} 
\\
  { \textsuperscript{1}Columbia University \ \
  \textsuperscript{2}UIUC \ \
  \textsuperscript{3}National University of Singapore 
  }
  \\
  \texttt{xudong.lin@columbia.edu} 
}
\maketitle

\begin{abstract}
Multi-channel video-language retrieval require models to understand information from different channels (e.g. video$+$question, video$+$speech)
to correctly link a video with a textual response or query. Fortunately, contrastive multimodal models are shown to be highly effective at aligning entities in images/videos and text, e.g., CLIP~\cite{radford2021learning}; text contrastive models are extensively studied recently for their strong ability of producing discriminative sentence embeddings, e.g., SimCSE~\cite{gao2021simcse}. However, there is not a clear way to quickly adapt these two lines to multi-channel video-language retrieval with limited data and resources. In this paper, we identify a principled model design space with two axes: how to represent videos and how to fuse video and text information. Based on categorization of recent methods, we investigate the options of representing videos using continuous feature vectors or discrete text tokens; for the fusion method, we explore the use of a multimodal transformer or a pretrained contrastive text model. We extensively evaluate the four combinations on five video-language datasets. We surprisingly find that discrete text tokens coupled with a pretrained contrastive text model yields the best performance, which can even outperform state-of-the-art on the iVQA and How2QA datasets without additional training on millions of video-text data. Further analysis shows that this is because representing videos as text tokens captures the key visual information and text tokens are naturally aligned with text models that are strong retrievers after the contrastive pretraining process. All the empirical analysis establishes a solid foundation for future research on affordable and upgradable multimodal intelligence.



  
\end{abstract}

\section{Introduction}

From retrieving a trending video on TikTok with natural language descriptions to asking a bot to solve your technical problem with the question and a descriptive video, AI agents handling multi-channel video-language retrieval-style tasks have been increasingly demanded in this post-social-media era. These tasks require the agent to fuse information from multiple channels, i.e., video and text to retrieve a text response or return a multi-channel sample for a text query.
To power such agents, a popular approach~\cite{li2020hero,luo2021clip4clip,li2022clip,yang2021just,zellers2021merlot} consists of two rounds of pretraining: (1) The 1st round is to obtain unimodal pretrained models, such as visual-only encoders~\cite{He_2016_CVPR,Kinetics,miech2019howto100m,radford2021learning} (e.g., S3D,CLIP) and text-only encoders~\cite{BERT,liu2019roberta,sanh2019distilbert,reimers-2019-sentence-bert} (e.g., BERT) (2) The 2nd round aims at pretraining on visual-text dataset - specifically, researchers leverage techniques like masked token modeling~\cite{li2020hero,zellers2021merlot} or contrastive learning~\cite{xu2021videoclip,luo2021clip4clip,li2022clip,yang2021just} to align and fuse unimodal features from model pretrained in the 1st round.

Such methods achieve good performance on multi-channel retrieval-style tasks but they suffer from two major limitations: 1) huge amounts of data and computational resources are required for the second-round ``pretraining'', which significantly limits the research exploration without such resources; 2) the domain of video data used in the second round ``pretraining'' has to be strongly correlated with downstream tasks~\cite{li2020hero}, which may restrict such methods from being generally applicable. 

To alleviate such limitations, we study a novel problem: fast adaptation of pretrained contrastive models on multi-channel video-language retrieval under limited resources. Specifically, we propose to adapt both contrastive multimodal models~\cite{miech2020end,radford2021learning} and contrastive text models~\cite{reimers-2019-sentence-bert,gao2021simcse} to enjoy their strong encoding ability and discriminative embedding space. There has been tremendous progress recently on large-scale contrastive multimodal models~\cite{miech2019howto100m,miech2020end,radford2021learning}. Through pretraining on millions of images/videos, these models are highly effective at \textit{encoding visual inputs and linking entities across modalities}. Meanwhile, contrastive text models~\cite{reimers-2019-sentence-bert,gao2021simcse} have been also densely studied to obtain \textit{discriminative sentence embeddings}. These models are shown to perform well on challenging text retrieval tasks such as semantic search~\cite{nguyen2016ms}, which requires the model to \textit{understand both language and real-world knowledge}~\cite{reimers-2019-sentence-bert, gao2021simcse}. 
Such an ability allows the model to retrieve a text response or encode a text query in multi-channel video-language retrieval-style tasks.
Thereby, once the video information is effectively incorporated, it eliminates the necessity of second-round ``pretraining'' on large scale multimodal datasets, enabling fast adaptation to any video-language retrieval-style tasks. 

\begin{figure*}[t!]
    \centering
    \includegraphics[width=0.99\textwidth]{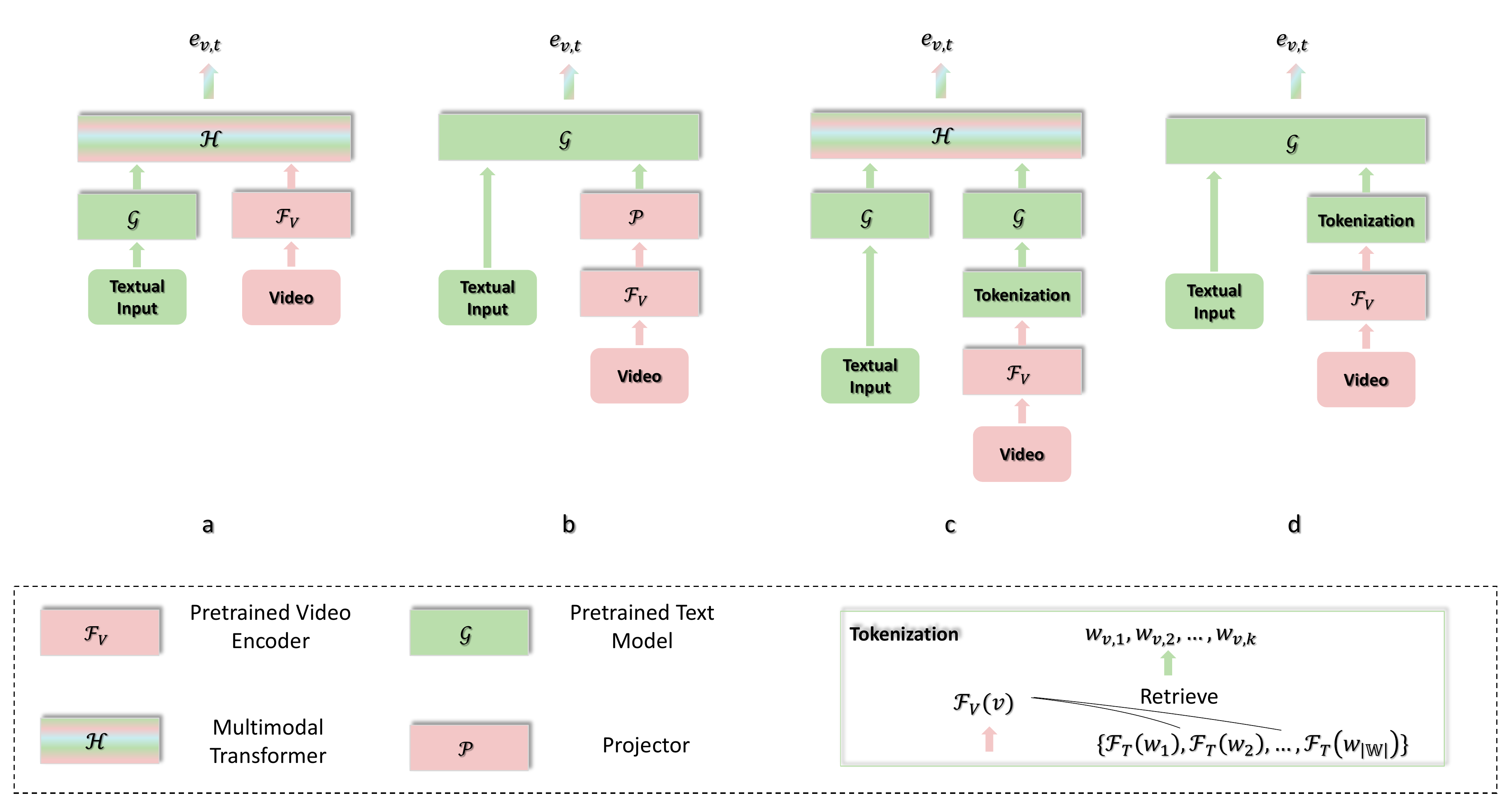}
    \caption{a. \textbf{Continuous Features + Multimodal Transformer}. This variant uses the pretrained video encoder to represent input video segment as continuous feature vectors. A randomly initialized multimodal transformer is used to fuse information of the video and textual input. 
    b. \textbf{Continuous Features + Text Transformer}. Instead of a multimodal transformer, we use the pretained contrastive text model to fuse projected video features and textual input.
    c. \textbf{Text Tokens + Multimodal Transformer}. It uses the pretrained video-text encoders to retrieve top-k text tokens from a predefined vocabulary to describe the video content. Then we feed the text tokens separately to the pretrained text model, to obtain contextualized embeddings for a multimodal transformer to fuse with contextualized text embeddings of other textual input. 
    d. \textbf{Text Tokens + Text Transformer}. We further remove the multimodal transformer and directly use the pretrained text model to fuse video and question. 
    }
    \label{illustration}
\end{figure*}

We first conduct a systematic analysis on potential model designs, as shown in Figure~\ref{illustration}. We identify the model design space with two design principles: how to represent the video information and how to fuse this video information with questions or other text such as speech. To represent video information, we could either adopt \textbf{Continuous Features} which is commonly used in existing work~\cite{mokady2021clipcap,yang2021just}, or project videos into unified \textbf{Text Tokens}~\cite{lin2021vx2text} from various modalities. To fuse information from multiple channels, i.e., video and question/speech, there are two potential options, namely, a \textbf{Multimodal Transformer}~\cite{yang2021just} or a \textbf{Text Transformer}~\cite{mokady2021clipcap}. 
Hence, there are four combinations derived from this model design space, namely, \textbf{Continuous Features + Multimodal Transformer}~\cite{yang2021just}, \textbf{Continuous Features + Text Transformer}, \textbf{Text Tokens + Multimodal Transformer}, \textbf{Text Tokens + Text Transformer}. 

Our exploration of this model design space results into a simple yet effective approach that allows fast adaptation of pretrained contrastive models, which first leverages contrastive multimodal models to retrieve a sequence of \textbf{Text Tokens} for the visual input and then feeds these tokens together with other text to contrastive \textbf{Text Transformer} for answer retrieval. Its fast adaptation ability not only comes from the ability of linking entities across modalities of the contrastive multimodal model, but it also enjoys the natural alignment with contrastive text models to produce discriminative embeddings. To the best of our knowledge, this is the first proposal to adapt pretrained contrastive models in this manner, although each individual design choice may have been adopted previously. 
We further conduct in-depth analysis to understand 1) the trade-off between data efficiency and accuracy, 2) the impact of pretrained contrastive text model, and 3) the possible limitation of this framework.

The contribution could be summarized three-fold:
\begin{itemize}
    \item We identified the principled model design space for fast adaption of pretrained contrastive multimodal models and pretrained contrastive text models.
    \item We conducted extensive experiments on five video-language datasets, observed a consistent trend across these four variants, and even obtained state-of-the-art performance (e.g., \textbf{6.5\%} improvement on How2QA) with the proposed Text Tokens + Text Transformer variant \textbf{without using millions of extra multimodal data samples}, which is essential to democratize the community of video+language.
    \item  The proposed Text Tokens + Text Transformer variant scales significantly better than the other variants, w.r.t. the quality of pretrained text models. The code will be released at \url{https://github.com/XudongLinthu/upgradable-multimodal-intelligence} to facilitate future research.  
\end{itemize}

\section{Related Work}
\label{rel}

\noindent \textbf{Pretrained Contrastive Models.} In this paper, we mainly consider two types of contrastive models: 1). contrastive multimodal models~\cite{radford2021learning,miech2020end,li2022clip,lin2022learning,wang2022object}, which typically consist of a visual encoder and a text encoder, and learn to map visual and text embeddings into a common space. They sample positive/negative pairs from aligned/unaligned image/video and text, and train the visual and text encoders with a contrastive objective in a self-supervised manner. With access to large-scale multimodal data (e.g., 400 million web image-text pairs~\cite{radford2021learning}), they are shown superior in linking entities across modalities; 2) Unlike general language models~\cite{BERT,T5}, contrastive text models~\cite{reimers-2019-sentence-bert,gao2021simcse}, which learn discriminative sentence embeddings from large-scale text data (e.g., 2 billion sentence pairs~\cite{n2019sbert}). They construct positive and negative sentence pairs and train the text encoder in a either supervised or unsupervised manner. 

\noindent \textbf{Multi-channel Vision-Language Learning.} Interests have been raised in multi-channel VL applications, where the model is asked to understand from both visual and language inputs to output a response (e.g., a phrase). Several works seek to get the response in a generative way~\cite{lin2021vx2text, le2019multimodal, zhou2020unified}, i.e., using a generative model to predict a sequence. On the other hand,  retrieval-style~\cite{yang2020bert, lu2016hierarchical, anderson2018bottom, VQA:ICCV2015} accomplishes the tasks in a discriminative way, where discriminative embeddings are produced to select the best answer from an answer candidate pool. In the paper, we focus on retrieval-style tasks, e.g., video question answering~\cite{yang2021just,yu2019activitynetqa,li2020hero,xu2017video} and multi-channel text-to-video retrieval~\cite{youcook2,wang2019vatex}.

\noindent \textbf{Adapting Pretrained Models to VL Applications. }
Existing works~\cite{tan2019lxmert,mokady2021clipcap,yang2021just,yang2022learning} explored different ways to adapt pretrained models to various vision-language tasks. For example, Shen \textit{et al.}~\cite{shen2021much} explored directly taking visual features from pretrained CLIP~\cite{radford2021learning} image encoders and feed them to existing vision-language models~\cite{tan2019lxmert} and then retrain the whole model using the new CLIP visual features. This empirical analysis shows that visual features learned by contrastive multimodal models are better for vision-language tasks. 
Just-ask~\cite{yang2021just} is a recent method for open-ended video question answering. 
The authors found the multimodal transformer gets significantly improved after a second-round pretraining on 69M video-question-answer triplets. All these methods require second-round pretraining on another large task-specific dataset, which requires additional engineering process and high demand of computational resources; on the other hand, we eliminate this process by efficiently leveraging the pretrained contrastive text models.

\section{Technical Approach}
\label{method}
In this section, we first include some preliminaries, and then introduce the four models in the model design space in detail. In the end, we describe the training technique and the implementation details.

\subsection{Preliminaries}
\textbf{Multi-channel Video-language Retrieval-style Tasks.}
These tasks require the model to fuse information from multiple channels, i.e., video and text to retrieve a text response or return a multi-channel sample for a text query. Specifically, we mainly consider open-ended video question answering and multi-channel text-video retrieval.
In open-ended video question answering, given a video $v$ and question $t$ as input, the model is required to retrieve the correct answer $a_k$ from a large answer corpus $\mathbb{A}=\{a_1,..,a_{|\mathbb{A}|}\}$. In multi-channel text-video retrieval, given a text query $a$, the model is required to retrieve the most relevant video $v_i$ with associated speech text $t_i$, which are from a corpus of multi-channel videos $\mathbb{V}=\{v_1,t_1,..,v_{|\mathbb{V}|},t_{|\mathbb{V}|}\}$. For simplicity, in the model descriptions, we will adopt open-ended video question answering for illustration.

\noindent \textbf{Pretrained Contrastive Multimodal Models.} We mainly leverage pretrained video-text contrastive models. It consists of a video encoder $\mathcal{F}_V: \mathbb{R}^{H \times W \times 3 \times F} \longrightarrow \mathbb{R}^{D}$ and a text encoder $\mathcal{F}_T: \mathbb{W}^{L} \longrightarrow \mathbb{R}^{D}$, where $H,W,F$ are the height, width and number of frames of the video, $L$ is the length of the sentence, $D$ is the dimension of the common embedding space and $\mathbb{W}$ is the set of all the words. Note that in all of the models, the pretrained contrastive multimodal models are \textbf{frozen} for fast adaptation.

\noindent \textbf{Pretrained Contrastive Text Models.}
The idea is to train a text encoder $\mathcal{G}: \mathbb{W}^{L} \longrightarrow \mathbb{R}^{D}$ to push the output embeddings of two relevant sentences to be similar and the irrelevant to be dissimilar. The pretrained contrastive text models will be \textbf{updated} in all of the models. Since all the actual instances of text models are transformer-based, text models and text transformers are used interchangeably.

\noindent \textbf{Multimodal Transformer and Projector.}
It is a shallow transformer model~\cite{vaswani2017attention} $\mathcal{H}: \mathbb{R}^{L \times D} \longrightarrow \mathbb{R}^{L \times D}$. We direct the audience for details of the transformer model to \cite{vaswani2017attention}. Inspired by prefix tuning~\cite{mokady2021clipcap}, we use the same transformer architecture for the projector $\mathcal{P}: \mathbb{R}^{L \times D} \longrightarrow \mathbb{R}^{L \times D}$, which is used to project video feature vectors to the input space of the pretrained text model.

\subsection{Model Variants in the Design Space}
For all the four variants, there is a separate encoder $\mathcal{G}_A$ to encode answers, which is initialized from pretrained text models. In the following, we mainly focus on the  difference of these four variants, i.e., the video representation and the multimodal fusion design.

\subsection*{a. \textbf{Continuous Features + Multimodal Transformer}}
As shown in Figure~\ref{illustration}, in this variant, we first directly use the pretrained video encoder $\mathcal{F}_V$ to represent input video segment $v$ as continuous feature vectors. The pretrained contrastive text model $\mathcal{G}$ is used to extract contextualized text embeddings from the question $t$. A randomly initialized multimodal transformer $\mathcal{H}$ is used to fuse information of the video and the question. Overall the model could be expressed as follows:
\begin{equation}
    e_{v,t} = \mathcal{H} (\mathcal{G} (t), \mathcal{F}_V(v) ),
\end{equation}
where $\mathcal{G} (t)$ are actually concatenated with  $\mathcal{F}_V(v)$ along the length axis before they are fed into the multimodal transformer.

\subsection*{b. \textbf{Continuous Features + Text Transformer}}
Inspired by prefix tuning work~\cite{mokady2021clipcap}, in this variant, we leverage a transformer projector $\mathcal{P}$ to project the continuous features from the video encoder $\mathcal{F}_V$ into the input space of the text transformer. Then the text transformer will fuse the information from both the video $v$ and the question $t$,
\begin{equation}
    e_{v,t} = \mathcal{G} ( t, \mathcal{P}(\mathcal{F}_V(v)) ),
\end{equation}
where the question $t$ actually first passes the embedding layer of $\mathcal{G}$ and then is concatenated with projected video features $\mathcal{P}(\mathcal{F}_V(v))$.

\subsection*{c. \textbf{Text Tokens + Multimodal Transformer}}
We propose the Text Token retrieval process as follows. First, we construct the word vectors from a predefined word vocabulary $\mathbb{W}$; specifically, for each word $w_i$, we use the text encoder to obtain its word vector $\mathcal{F}_T(w_i)$. For the input video $v$, we encode it with the pretrained video encoder $\mathcal{F}_V(v)$. Then we compare the similarity between the video and each of the words to retrieve $k$ most similar words,
\begin{equation}
    w_{v,1},.., w_{v,k}  =  \arg \max_i^k  \mathcal{F}_T(w_i)^\top \mathcal{F}_V(v).
\end{equation}

Then the retrieved words are fed into the pretrained text model with the question in parallel to obtain the contextualized embeddings, which are finally concatenated to be fed into the multimodal transformer,
\begin{equation}
    e_{v,t} = \mathcal{H} (\mathcal{G} (t), \mathcal{G} (w_{v,1},.., w_{v,k}) ).
\end{equation}

Note that to maintain similar parameter usage, we share the weights between the text model $\mathcal{G}$ taking question as input and the text model $\mathcal{G}$ taking video words as input.

\subsection*{d. \textbf{Text Tokens + Text Transformer}}
In this variant, we follow the same text token retrieval process and then further simplify the model to use pretrained text model for modality fusion. Formally, it can be described as,
\begin{equation}
    e_{v,t} = \mathcal{G} (t, w_{v,1},.., w_{v,k}),
\end{equation}
where the question $t$ and all the retrieved words are concatenated as a whole text sequence and then input to the pretrained text model.
The removal of the multimodal transformer enables this variant to directly benefit from the pretrained discriminative sentence embedding space of $\mathcal{G}$.

\subsection{Training}
During training, we leverage the NCE loss~\cite{yang2021just} to optimize the parameters $\theta$ of $\mathcal{G}$, $\mathcal{G_A}$ $\mathcal{H}$, and $\mathcal{P}$ (if included in the model). In the actual training process, we calculate loss in a mini-batch manner. For simplicity, we omit average over a batch in the following equations.

For open-ended video question answering, we optimize the following objective
\begin{equation} 
    \min _\theta - \log \frac{\exp \left( e_{v,t} ^\top \mathcal{G}_A (a_l) \right)}{\sum_i \exp \left( e_{v,t}^\top \mathcal{G}_A (a_i) \right) },
\end{equation}
where $l$ is the ground-truth answer index. 

For multi-channel text-video retrieval, we adopt the symmetric version of the loss~\cite{radford2021learning}, Specifically,  
\begin{equation}
\begin{aligned} 
    \min _\theta &-\frac{1}{2}( \log \frac{\exp \left( e_{v,t} ^\top \mathcal{G}_A (a_l) \right)}{\sum_i \exp \left( e_{v,t}^\top \mathcal{G}_A (a_i) \right) } \\
    &+     
    \log \frac{\exp \left( e_{v,t} ^\top \mathcal{G}_A (a_l) \right)}{\sum_i \exp \left( e_{v_i,t_i}^\top \mathcal{G}_A (a_l) \right) }),
\end{aligned}
\end{equation}
where $e_{v,t}$ and $a_l$ are a pair of embeddings of multi-channel video and text query, and $e_{v_i,t_i}$ is the representation of the i-th multi-channel video in the mini-batch. Note that although $\mathcal{G_A}$ is also initialized from the pretrained text model, it does not share parameters with $\mathcal{G}$ during training, following the setting of~\cite{yang2021just}.

 \begin{table*}[h]
\centering

\fontsize{9pt}{10pt} \selectfont
\begin{tabu}{cccccc}
\toprule
\multirow{2}{*}{Model} & \multicolumn{2}{c}{Open-ended Acc(\%)} & Multi-Choice Acc(\%) & \multicolumn{2}{c}{Retrieval AveR (\%)} \\
   & iVQA   & ActivityNet-QA  & How2QA    & YouCook II & VATEX
\\ \midrule

 Conti. + Multi. & 22.4  & 36.9   & 79.2  & 41.9 & 69.4 \\
 Conti. + Text & 23.2   & 37.3  & 80.4 & 46.2 & 72.7\\
  Text + Multi. & 23.4  & 37.1    & 79.4  & 40.4 & 67.5\\
 \textbf{Text + Text}    & \textbf{31.6}  & \textbf{38.7}    & \textbf{82.9}  & \textbf{49.7} & \textbf{74.8}\\
\bottomrule
\end{tabu}

\caption{Comparison of the four model variants on various multi-channel video-text retrieval-style tasks. Acc is short for accuracy. AveR is short for average recall@\{1,5,10\}.
}
\label{ex1}
\end{table*}

\subsection{Implementation Details}
We use MPNet (all-mpnet-base-v2)~\cite{mpnet,n2019sbert} as $\mathcal{G}$ when not specified. It is ranked first by Sentence Transformers~\cite{n2019sbert} at the time of writing (November, 2022).
We use the pretrained models from \cite{miech2020end} for $\mathcal{F}_T$ and $\mathcal{F}_V$ when not specified. When using CLIP~\cite{radford2021learning}, we follow FrozenBiLM~\cite{yang2022zero}.
The resulted video feature vector is at 1 vector per 1.5 seconds.
For $\mathcal{H}$ and $\mathcal{P}$, we use the same shallow multimodal transformer as in~\cite{yang2021just}, which has 2 transformer blocks with 512/768 as the embedding dimension. We follow the training procedure and hyper-parameter settings for all the variants to conduct fair comparisons. Details are provided in the supplementary material\footnote{\url{https://arxiv.org/abs/2206.02082}}.

We find that the vocabulary $\mathbb{W}$ can be effectively constructed by parsing all the query/answer sentences in the downstream datasets with spaCy~\cite{honnibal2017spacy} to obtain a set of unique verbs and nouns.
We use $k=15$ words retrieved from the vocabulary for each feature vector in the video. We follow \cite{yang2021just} to sub-sample video feature vectors when the video is too long to fit the memory. We further apply a max pooling with kernel size as 5 to sub-sample the retrieved words to avoid too many repetitions of words of neighbouring segmenting.




\section{Experimental Results}
In this section, we will first introduce the datasets and evaluation metric, then we organize the following subsections by answering a set of important questions of the four model variants. Extra ablations are provided in the supplementary material for hyper-parameter selection. 

\subsection{Dataset and Evaluation Metric} 
We select 5 commonly used multi-channel retrieval-style datasets: \textbf{iVQA~\cite{yang2021just}}, \textbf{How2QA~\cite{li2020hero}}, \textbf{ActivityNet-QA~\cite{yu2019activitynetqa}}, \textbf{YouCook II~\cite{ZhXuCoAAAI18}}, and \textbf{VATEX~\cite{wang2019vatex}} as the main evaluation datasets. For the first three datasets, accuracy is used as the evaluation metric and for the other two, an average of recall@\{1,5,10\} is used.  Details are provided in the supplementary material.

\noindent \textbf{MSRVTT-QA and MSVD-QA~\cite{xu2017video}}. These two datasets are automatically generated from video-caption pairs. The data is too noisy compared to the aforementioned manually annotated datasets but this actually could serve as a testbed to see under what circumstances our variants won't work well. We follow~\cite{yang2021just} for experimental settings and evaluation metric on these two datasets.

\begin{figure}[t]
    \centering
    \includegraphics[width=0.45\textwidth]{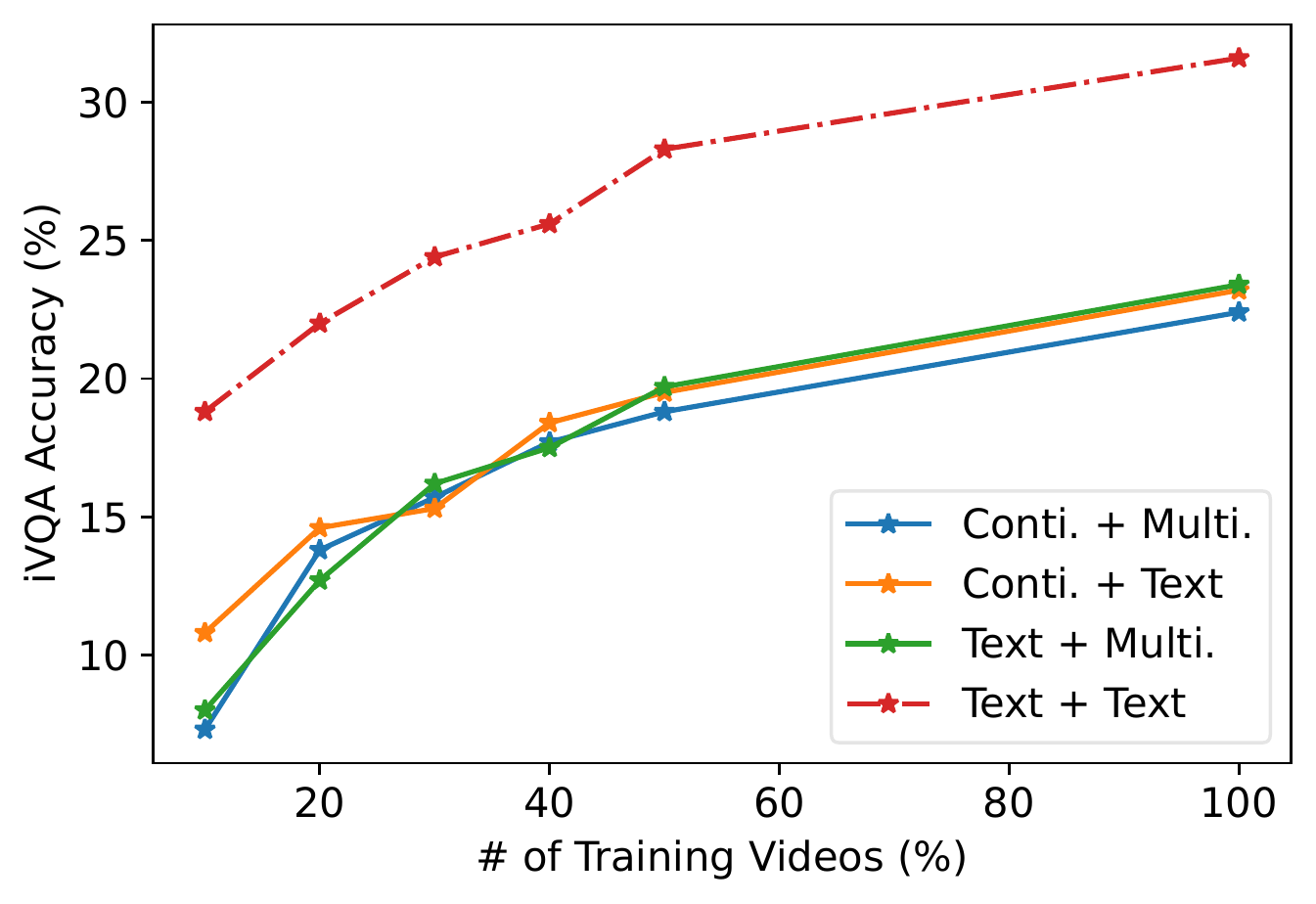}
    
    \caption{Comparison of the four variants under few-shot setting on the iVQA dataset.}
    \label{fig:few}
\end{figure}

\subsection{Which of the variants performs the best?}

As shown in Table~\ref{ex1}, the \textbf{Text Tokens + Text Transformer} consistently performs better than the other three variants when we directly tune these models without second-round large-scale multimodal ``pretraining''.
We suppose this is because of this variant can easily benefit from both pretrained contrastive multimodal models and contrastive text models without struggling to align the space of them or training a multimodal transformer from scratch. We also observe that on downstream task from similar domains with that of the pretrained contrastive multimodal model~\cite{miech2020end}, our proposed variant usually enjoys higher improvement, e.g., on iVQA and YouCook II. 

\begin{table*}[t]

\centering

\fontsize{9pt}{10pt} \selectfont
\begin{tabu}{ccccccc}

\toprule
Model & $\mathcal{F}_V, \mathcal{F}_T$ &  Extra MM Samples   & $\Delta$ GPU hours & iVQA  & ActivityNet  & How2QA
\\ \midrule
\rowfont{\color{gray}} MERLOT~\cite{zellers2021merlot} & - & 180M & - & -& 41.4 & -\\
\rowfont{\color{gray}} SiaSamRea~\cite{yu2021learning} & - &5.6M + 80K        & - & - & 39.8 & 84.1\\
\midrule

VQA-T~\cite{yang2022learning} & S3D~\cite{miech2020end} & 69M + 3M       & 350 + 30 & 35.2 & \textbf{39.0} & \textbf{85.3}\\
 Conti. + Multi. & S3D~\cite{miech2020end} & 69M        & 400 & 35.4 & 38.9 & 84.4\\
 Conti. + Multi. (\textit{+ ASR}) & S3D~\cite{miech2020end} & 69M        & 400 & 36.0 & 38.9 & 84.8\\
 Text + Text (Ours)  & S3D~\cite{miech2020end} & 0 & 0   & 31.6     & 38.7 & 82.9 \\
 Text + Text (\textit{+ ASR}, Ours)  & S3D~\cite{miech2020end} & 0        & 0& \textbf{36.8} & 38.8 & 84.6\\
 \midrule 
  FrozenBiLM~\cite{yang2022zero} & CLIP~\cite{radford2021learning} & 10M & 160 & 39.7 & 43.2 & 81.5 \\
 FrozenBiLM~\cite{yang2022zero} (\textit{+ ASR}) & CLIP~\cite{radford2021learning} & 10M & 160 & 39.6 & \textbf{43.2} & 86.7 \\
  Text + Text (Ours)  & CLIP~\cite{radford2021learning} & 0 & 0   & 36.9     & 41.4 & 92.4 \\
 Text + Text (\textit{+ ASR}, Ours) & CLIP~\cite{radford2021learning} & 0        & 0& \textbf{40.2} & 41.4 & \textbf{93.2} \\

\bottomrule
\end{tabu}

\caption{Comparison with the state-of-the-art on iVQA, ActivityNet and How2VQA in terms of accuracy and efficiency. Extra MM Samples indicate the number of video-text samples that are needed in the second-round pretraining. $\Delta$ GPU hours refer to the additional computation required for the second-round pretraining. \textit{+ ASR} indicates the use of ASR texts of the video as additional inputs. Note that our variant typically requires 0.5 GPU hours for training. Methods in gray enjoy a more costly end-to-end training process.
}
\label{ex2}
\end{table*}

To further understand the behavior of these four variants with limited training data for downstream adaptation, we explore the few-shot setting. We sample a subset of iVQA and train the four variants with the same sub-sampled set for the same number of iterations with that of the full-shot setting. In Figure~\ref{fig:few}, we consistently observe a large margin between \textbf{Text Tokens + Text Transformer} and the other variants. 
Continuous Features + Multimodal Transformer generally performs the worst on the iVQA dataset and also other datasets, which implicitly verifies our hypothesis that it is crucial to directly leverage the aligned representation space and the discriminative output space. 



\subsection{Comparison with state-of-the-art on accuracy-efficiency trade-off}
We aim to examine the trade-off between downstream performance and the additional resource requirements, through comparison between our best variant with state-of-the-art models that leverage huge amounts of data and computational resources for pretraining the multimodal transformer~\cite{yang2022learning,yang2022zero}.
As shown in Table~\ref{ex2}, with \textbf{three orders of magnitude smaller amount GPU training time}, our model can actually achieve comparable or even better results than the state-of-the-art. When using pretrained $\mathcal{F}_V, \mathcal{F}_T$ from~\cite{miech2020end}, our model can even outperform the state-of-the-art by 0.8\% when with ASR as input. 

When switching the $\mathcal{F}_V, \mathcal{F}_T$ to CLIP~\cite{radford2021learning}, our \textbf{Text Tokens + Text Transformer} even significantly outperforms than models that enjoys a more costly end-to-end second-round ``pretraining'' process like MERLOT~\cite{zellers2021merlot} or SiaSamRea~\cite{yu2021learning}. Most encouragingly, our model also outperforms the state-of-the-art FrozenBiLM on How2QA ($\textbf{+6.5\%}$) and iVQA ($\textbf{+0.5\%}$), which uses 10 million samples for second-round ``pretraining'' and uses a much larger text model.

We also would like to highlight that our \textbf{Text Tokens + Text Transformer} \textit{easily benefits from a better pretrained $\mathcal{F}_V, \mathcal{F}_T$} as we observe significant improvement for our model, which is encouraging for future development of our proposed model when better pretrained multimodal contrastive models are available. Comparisons on the other two datasets are separately provided in the supplementary material as the compared methods in Table~\ref{ex2} are not state-of-the-art on them.


\subsection{How well does \textbf{Text Tokens + Text Transformer} benefit from pretrained text models?}
\begin{figure*}[t]
    \centering
    \includegraphics[width=0.9\linewidth]{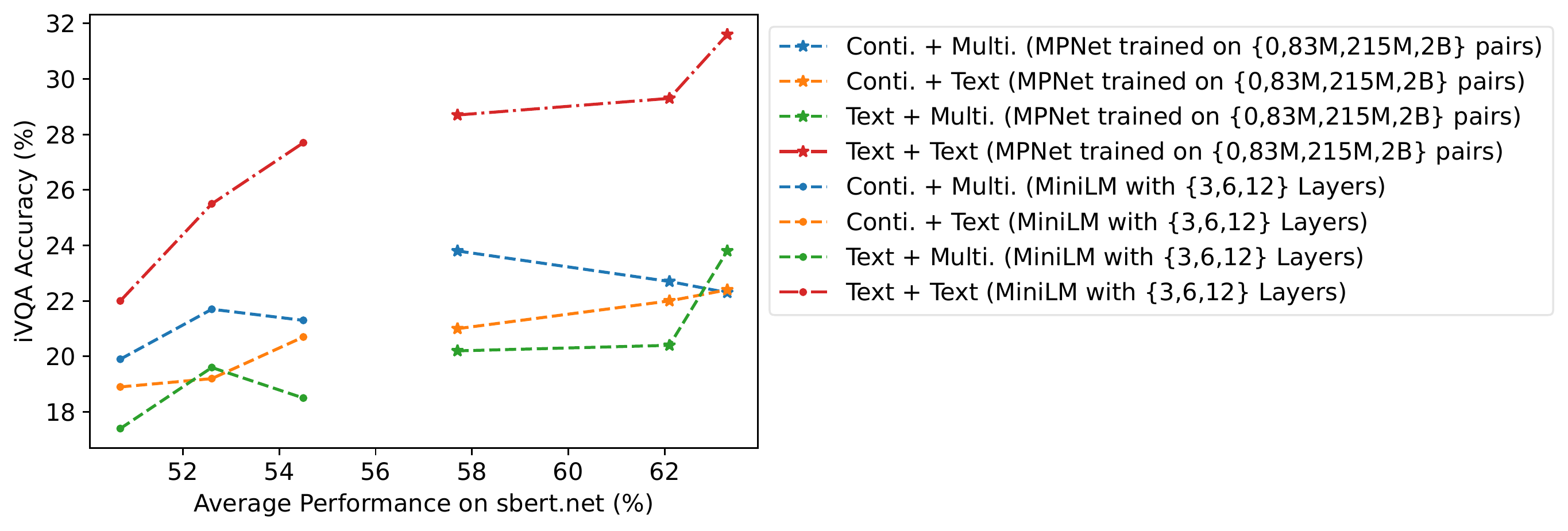}
    \caption{Results of varying the pretrained text model in the four variants on the iVQA dataset. The performance of all SBERT models are obtained from~\cite{n2019sbert}.}
    \label{fig:sbert}
\end{figure*}

\begin{figure}[!htb]
    \centering
        \includegraphics[width=0.85\linewidth]{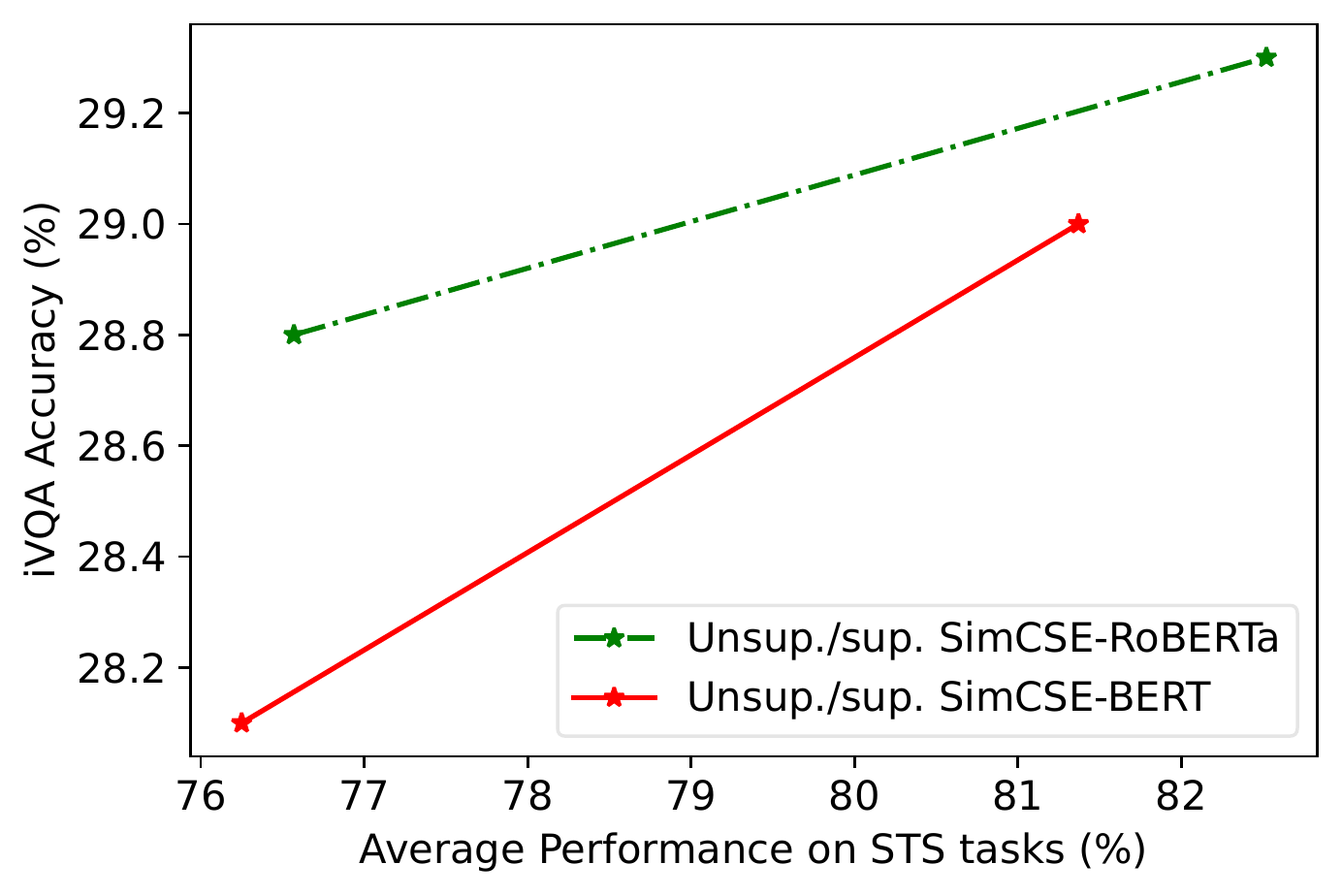}
        
    \caption{ Results of varying the pretrained text model in the \textbf{Text Tokens + Text Transformer} on the iVQA dataset. The semantic textual
similarity (STS) performance of SimCSE models is obtained from~\cite{gao2021simcse}. }
    \label{fig:simcse}
\end{figure}

To comprehensively understand the effect of pretrained contrastive text models, we investigate two sources of models, as shown in Figure~\ref{fig:sbert} and Figure~\ref{fig:simcse}. SBERT~\cite{n2019sbert} is a widely-used library with models trained on paired sentences. We measure the quality of pretrained text models with the average performance over 20 sentence embedding/semantic search tasks. SimCSE~\cite{gao2021simcse} explores unsupervised contrastive learning of transformer models and provides various models. We measure the quality of pretrained text models with the average performance over 7 semantic textual similarity (STS) tasks.

In Figure~\ref{fig:sbert}, we first study the four variants trained on the same data with different model sizes. MiniLM~\cite{wang2020minilm} (paraphrase-MiniLM~\cite{n2019sbert}) with $\{3,6,12\}$ layers is a suitable testbed as the model architecture is identical but the number of layers changes.  We obtain a clear positive correlation between the quality of text models and the accuracy on iVQA. We then study the same model trained with different size of data: paraphrase-mpnet-base-v2, multi-qa-mpnet-base-dot-v1, and all-mpnet-base-v2, which are trained on 83 million, 215 million, and 2 billion sentence pairs, respectively. Again our proposed variant scales well with the performance of the text models. 

The scalability w.r.t. the quality of pretrained text model of the proposed \textbf{Text Tokens + Text Transformer} is surprisingly superior compared the other variants. As shown in Figure~\ref{fig:sbert}, among all the four variants, the proposed \textbf{Text Tokens + Text Transformer} has the most positive correlation between the multimodal task performance on iVQA and the quality of the pretrained text models. We also observe that both models using pretrained text transformer for multimodal fusion have a positive correlation and both models using multimodal transformer for multimodal fusion don't benefit from the improvement of the pretrained text models. From this comparison, we conclude that only the proposed \textbf{Text Tokens + Text Transformer} can well enjoy the improvement of pretrained text models.

We also verify the necessity of using pretrained contrastive text models by directly tuning the MPNet from~\cite{mpnet} without contrastive pretraining. It is not shown in Figure~\ref{fig:sbert} as its text performance is not evaluated by sbert.net. This model only achieves $22.0\%$, which is significantly lower than the three contrastively pretrained versions.

Then we compare both the unsupervised and supervised BERT/RoBERTa models from SimCSE. As shown in Figure~\ref{fig:simcse}, as STS performance increases, the overall downstream performance on iVQA is improved, despite incoherence across different model architectures. As a summary, even with various model architectures, training methods, supervision sources and dataset sizes, we consistently observe positive correlation, which is highly encouraging for further upgradability of the proposed \textbf{Text Tokens + Text Transformer} in the future. 



\subsection{Model Interpretation}
To understand why our proposed \textbf{Text Tokens + Text Transformer} performs well, we select iVQA to first check a simple statistic: the proportion of test samples that have at least one word overlapped between the answer and retrieved text tokens of the video. We find that the proportion is actually $66.4\%$, which partially explains why we obtain a huge improvement on the iVQA dataset. 

\begin{figure*}[!htb]
    \centering
    \begin{subfigure}{.99\textwidth}
        \centering
        \includegraphics[width=0.99\linewidth, trim={0 0.6cm 0 0.6cm},clip]{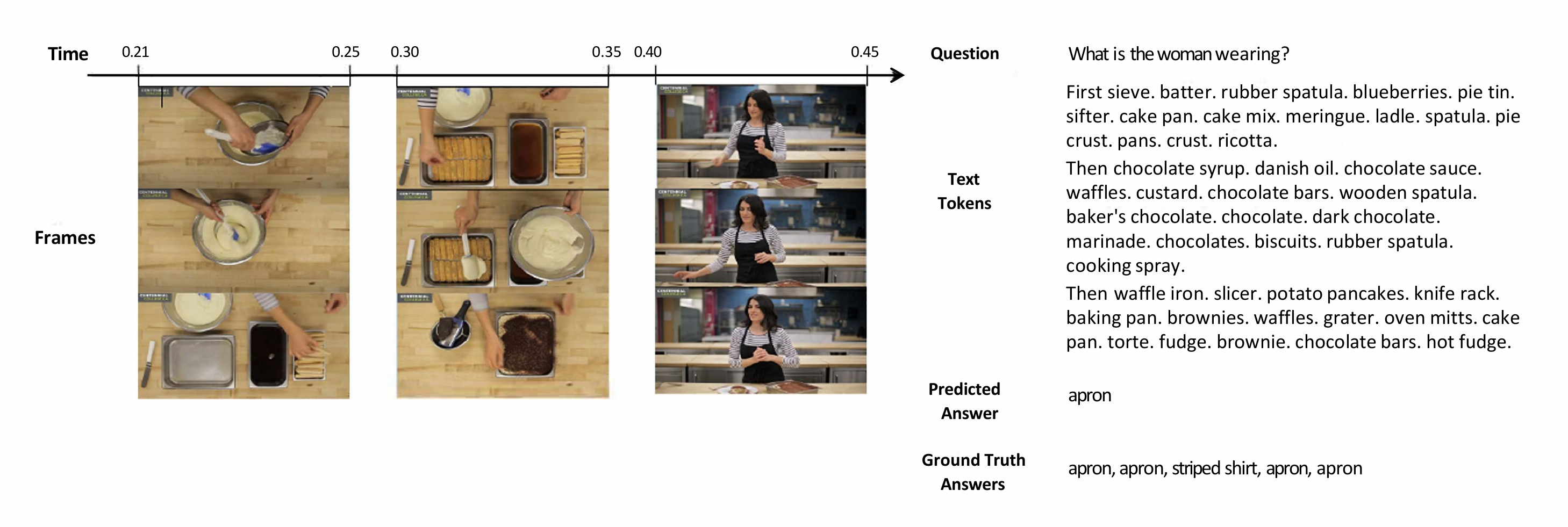}
        \caption{Successful case.}
        \label{fig:suc}
    \end{subfigure}
    \\
    \begin{subfigure}{0.99\textwidth}
        \centering
        \includegraphics[width=0.99\linewidth, trim={0 0.9cm 0 0.6cm},clip]{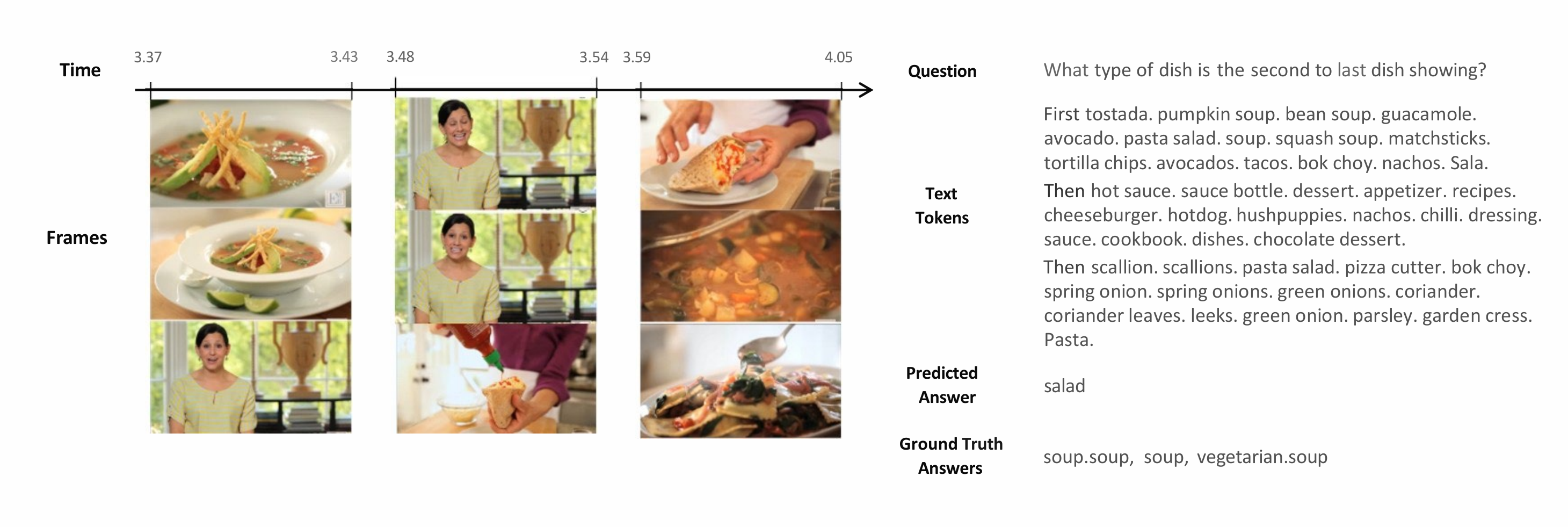}
        \caption{Failure case}
        \label{fig:fail}
    \end{subfigure}
    \caption{ Visualization of \textbf{Text Tokens + Text Transformer} on iVQA. ``First'' and ``then'' indicate the temporal order and they have minimal effect on performance ($<0.5\%$). 
    }
    \label{fig:vis}
\end{figure*}

We visualize one successful case and one failure case in Figure~\ref{fig:vis}. In Figure~\ref{fig:suc}, the model is not able to retrieve the answer word ``apron'' for the video but based on the rich kitchen-related context words, the text-transformer can still answer the question correctly. This indicates that our proposed model can accommodate imperfection of the tokenization process.
In Figure~\ref{fig:fail}, although ``soup'' is retrieved as a text token for the video in the first few segments of the video, the crucial video segment required fine-grained temporal understanding of the video in the last segment and the text tokens after max pooling is dominated by other food-related words. 
But overall, according the aforementioned statistic, this Text Tokens + Text Transformer approach certainly enjoys the high explainability in its design, compared with other using Continuous Features.

\subsection{Results on Automatically Generated Datasets}
We evaluate the proposed \textbf{Text Tokens + Text Transformer} on the automatically generated MSRVTT-QA and MSVD-QA datasets. We find that Text Tokens + Text Transformer (39.7\%, 40.9\%) performs similarly with the Continuous Features +  Multimodal Transformer baseline (39.6\%, 41.2\%). To quantify the imperfection of question generation, we carefully convey a manual study on 50 randomly sampled question-answer-video triplets in the MSRVTT-QA dataset. We found that the 6\% of the questions are not exactly aligned with the video, e.g., wrong entity descriptions/wrong action descriptions. 24\% of the questions have grammatical errors. 10\% of the questions are ambiguous. 
These results indicate that when the text queries have very low-quality, the well-trained text embedding space may suffer from the style change of the input language. 
More discussion is included in the supplementary.



\section{Conclusion}
We aim at a novel and challenging problem: fast adaption of pretrained contrastive multimodal models and pretrained contrastive text models for multi-channel video-language retrieval under limited data and computation resources. We systematically evaluate four variants of models from our identified principled model design space. Through extensive analysis, we find the variant with text tokens as video representations and contrastive text model for multimodal fusion achieve the best performance. Without millions of multimodal data for pretraining and three orders of magnitude more training time, our model achieves comparable performance with the state-of-the-art model. We further show that this simple yet effective variant can be easily improved by using better pretrained contrastive multimodal models or pretrained contrastive text models, which uncovers the great potential of our proposed model to democratize the video-language research community from heavy dependence on huge data and computation resources towards upgradable multimodal intelligence. 

\section{Acknowledgement}
We thank the anonymous reviewers, Richard Zemel, other DVMMers and colleagues for their feedback. This paper is based upon work supported by U.S. DARPA KAIROS Program No. FA8750-19-2-1004. The views and conclusions contained herein are those of the authors and should not be interpreted as necessarily representing the official policies, either expressed or implied, of DARPA, or the U.S. Government. The U.S. Government is authorized to reproduce and distribute reprints for governmental purposes notwithstanding any copyright annotation therein. Mike Zheng Shou does not receive any funding for this work.

{\small
\bibliographystyle{ieee_fullname}
\bibliography{egbib}

\begin{thebibliography}{10}\itemsep=-1pt

\bibitem{anderson2018bottom}
Peter Anderson, Xiaodong He, Chris Buehler, Damien Teney, Mark Johnson, Stephen
  Gould, and Lei Zhang.
\newblock Bottom-up and top-down attention for image captioning and visual
  question answering.
\newblock In {\em Proceedings of the IEEE conference on computer vision and
  pattern recognition}, pages 6077--6086, 2018.

\bibitem{VQA:ICCV2015}
Stanislaw Antol, Aishwarya Agrawal, Jiasen Lu, Margaret Mitchell, Dhruv Batra,
  C Lawrence~Zitnick, and Devi Parikh.
\newblock Vqa: Visual question answering.
\newblock In {\em Proceedings of the IEEE international conference on computer
  vision}, pages 2425--2433, 2015.

\bibitem{ba2016layer}
Jimmy~Lei Ba, Jamie~Ryan Kiros, and Geoffrey~E Hinton.
\newblock Layer normalization.
\newblock {\em arXiv preprint arXiv:1607.06450}, 2016.

\bibitem{s2016activitynet}
Fabian Caba~Heilbron, Victor Escorcia, Bernard Ghanem, and Juan Carlos~Niebles.
\newblock Activitynet: A large-scale video benchmark for human activity
  understanding.
\newblock In {\em Proceedings of the ieee conference on computer vision and
  pattern recognition}, pages 961--970, 2015.

\bibitem{Kinetics}
Joao Carreira and Andrew Zisserman.
\newblock Quo vadis, action recognition? a new model and the kinetics dataset.
\newblock In {\em CVPR}, 2017.

\bibitem{BERT}
Jacob Devlin, Ming-Wei Chang, Kenton Lee, and Kristina Toutanova.
\newblock Bert: Pre-training of deep bidirectional transformers for language
  understanding.
\newblock {\em arXiv preprint arXiv:1810.04805}, 2018.

\bibitem{feichtenhofer2019slowfast}
Christoph Feichtenhofer, Haoqi Fan, Jitendra Malik, and Kaiming He.
\newblock Slowfast networks for video recognition.
\newblock In {\em Proceedings of the IEEE/CVF international conference on
  computer vision}, pages 6202--6211, 2019.

\bibitem{gao2021simcse}
Tianyu Gao, Xingcheng Yao, and Danqi Chen.
\newblock Simcse: Simple contrastive learning of sentence embeddings.
\newblock {\em arXiv preprint arXiv:2104.08821}, 2021.

\bibitem{He_2016_CVPR}
Kaiming He, Xiangyu Zhang, Shaoqing Ren, and Jian Sun.
\newblock Deep residual learning for image recognition.
\newblock In {\em CVPR}, 2016.

\bibitem{heilman2010good}
Michael Heilman and Noah~A Smith.
\newblock Good question! statistical ranking for question generation.
\newblock In {\em Human Language Technologies: The 2010 Annual Conference of
  the North American Chapter of the Association for Computational Linguistics},
  pages 609--617, 2010.

\bibitem{honnibal2017spacy}
Matthew Honnibal and Ines Montani.
\newblock spacy 2: Natural language understanding with bloom embeddings,
  convolutional neural networks and incremental parsing.
\newblock {\em To appear}, 7(1):411--420, 2017.

\bibitem{le2019multimodal}
Hung Le, Doyen Sahoo, Nancy Chen, and Steven Hoi.
\newblock Multimodal transformer networks for end-to-end video-grounded
  dialogue systems.
\newblock In {\em Proceedings of the 57th Annual Meeting of the Association for
  Computational Linguistics}, pages 5612--5623, 2019.

\bibitem{li2020hero}
Linjie Li, Yen-Chun Chen, Yu Cheng, Zhe Gan, Licheng Yu, and Jingjing Liu.
\newblock Hero: Hierarchical encoder for video+ language omni-representation
  pre-training.
\newblock {\em arXiv preprint arXiv:2005.00200}, 2020.

\bibitem{li2021value}
Linjie Li, Jie Lei, Zhe Gan, Licheng Yu, Yen-Chun Chen, Rohit Pillai, Yu Cheng,
  Luowei Zhou, Xin~Eric Wang, William~Yang Wang, et~al.
\newblock Value: A multi-task benchmark for video-and-language understanding
  evaluation.
\newblock {\em arXiv preprint arXiv:2106.04632}, 2021.

\bibitem{li2022clip}
Manling Li, Ruochen Xu, Shuohang Wang, Luowei Zhou, Xudong Lin, Chenguang Zhu,
  Michael Zeng, Heng Ji, and Shih-Fu Chang.
\newblock Clip-event: Connecting text and images with event structures.
\newblock {\em arXiv preprint arXiv:2201.05078}, 2022.

\bibitem{lin2021vx2text}
Xudong Lin, Gedas Bertasius, Jue Wang, Shih-Fu Chang, Devi Parikh, and Lorenzo
  Torresani.
\newblock Vx2text: End-to-end learning of video-based text generation from
  multimodal inputs.
\newblock In {\em Proceedings of the IEEE/CVF Conference on Computer Vision and
  Pattern Recognition}, pages 7005--7015, 2021.

\bibitem{lin2022learning}
Xudong Lin, Fabio Petroni, Gedas Bertasius, Marcus Rohrbach, Shih-Fu Chang, and
  Lorenzo Torresani.
\newblock Learning to recognize procedural activities with distant supervision.
\newblock In {\em Proceedings of the IEEE/CVF Conference on Computer Vision and
  Pattern Recognition}, pages 13853--13863, 2022.

\bibitem{liu2019roberta}
Yinhan Liu, Myle Ott, Naman Goyal, Jingfei Du, Mandar Joshi, Danqi Chen, Omer
  Levy, Mike Lewis, Luke Zettlemoyer, and Veselin Stoyanov.
\newblock Roberta: A robustly optimized bert pretraining approach.
\newblock {\em arXiv preprint arXiv:1907.11692}, 2019.

\bibitem{lu2016hierarchical}
Jiasen Lu, Jianwei Yang, Dhruv Batra, and Devi Parikh.
\newblock Hierarchical question-image co-attention for visual question
  answering.
\newblock In {\em Advances in neural information processing systems}, pages
  289--297, 2016.

\bibitem{luo2021clip4clip}
Huaishao Luo, Lei Ji, Ming Zhong, Yang Chen, Wen Lei, Nan Duan, and Tianrui Li.
\newblock Clip4clip: An empirical study of clip for end to end video clip
  retrieval.
\newblock {\em arXiv preprint arXiv:2104.08860}, 2021.

\bibitem{miech2020end}
Antoine Miech, Jean-Baptiste Alayrac, Lucas Smaira, Ivan Laptev, Josef Sivic,
  and Andrew Zisserman.
\newblock End-to-end learning of visual representations from uncurated
  instructional videos.
\newblock In {\em Proceedings of the IEEE/CVF Conference on Computer Vision and
  Pattern Recognition}, pages 9879--9889, 2020.

\bibitem{miech2019howto100m}
Antoine Miech, Dimitri Zhukov, Jean-Baptiste Alayrac, Makarand Tapaswi, Ivan
  Laptev, and Josef Sivic.
\newblock Howto100m: Learning a text-video embedding by watching hundred
  million narrated video clips.
\newblock In {\em Proceedings of the IEEE/CVF International Conference on
  Computer Vision}, pages 2630--2640, 2019.

\bibitem{mokady2021clipcap}
Ron Mokady, Amir Hertz, and Amit~H Bermano.
\newblock Clipcap: Clip prefix for image captioning.
\newblock {\em arXiv preprint arXiv:2111.09734}, 2021.

\bibitem{nguyen2016ms}
Tri Nguyen, Mir Rosenberg, Xia Song, Jianfeng Gao, Saurabh Tiwary, Rangan
  Majumder, and Li Deng.
\newblock Ms marco: A human generated machine reading comprehension dataset.
\newblock In {\em CoCo@ NIPS}, 2016.

\bibitem{ni2021large}
Jianmo Ni, Chen Qu, Jing Lu, Zhuyun Dai, Gustavo~Hern{\'a}ndez {\'A}brego, Ji
  Ma, Vincent~Y Zhao, Yi Luan, Keith~B Hall, Ming-Wei Chang, et~al.
\newblock Large dual encoders are generalizable retrievers.
\newblock {\em arXiv preprint arXiv:2112.07899}, 2021.

\bibitem{radford2021learning}
Alec Radford, Jong~Wook Kim, Chris Hallacy, Aditya Ramesh, Gabriel Goh,
  Sandhini Agarwal, Girish Sastry, Amanda Askell, Pamela Mishkin, Jack Clark,
  et~al.
\newblock Learning transferable visual models from natural language
  supervision.
\newblock In {\em International Conference on Machine Learning}, pages
  8748--8763. PMLR, 2021.

\bibitem{T5}
Colin Raffel, Noam Shazeer, Adam Roberts, Katherine Lee, Sharan Narang, Michael
  Matena, Yanqi Zhou, Wei Li, and Peter~J Liu.
\newblock Exploring the limits of transfer learning with a unified text-to-text
  transformer.
\newblock {\em arXiv preprint arXiv:1910.10683}, 2019.

\bibitem{reimers-2019-sentence-bert}
Nils Reimers and Iryna Gurevych.
\newblock Sentence-bert: Sentence embeddings using siamese bert-networks.
\newblock In {\em Proceedings of the 2019 Conference on Empirical Methods in
  Natural Language Processing}. Association for Computational Linguistics, 11
  2019.

\bibitem{n2019sbert}
Nils Reimers and Iryna Gurevych.
\newblock Sentence {T}ransformers.
\newblock \url{https://www.sbert.net/}, 2019.

\bibitem{sanh2019distilbert}
Victor Sanh, Lysandre Debut, Julien Chaumond, and Thomas Wolf.
\newblock Distilbert, a distilled version of bert: smaller, faster, cheaper and
  lighter.
\newblock {\em arXiv preprint arXiv:1910.01108}, 2019.

\bibitem{shen2021much}
Sheng Shen, Liunian~Harold Li, Hao Tan, Mohit Bansal, Anna Rohrbach, Kai-Wei
  Chang, Zhewei Yao, and Kurt Keutzer.
\newblock How much can clip benefit vision-and-language tasks?
\newblock {\em arXiv preprint arXiv:2107.06383}, 2021.

\bibitem{mpnet}
Kaitao Song, Xu Tan, Tao Qin, Jianfeng Lu, and Tie-Yan Liu.
\newblock Mpnet: Masked and permuted pre-training for language understanding.
\newblock In H. Larochelle, M. Ranzato, R. Hadsell, M.~F. Balcan, and H. Lin,
  editors, {\em Advances in Neural Information Processing Systems}, volume~33,
  pages 16857--16867. Curran Associates, Inc., 2020.

\bibitem{tan2019lxmert}
Hao Tan and Mohit Bansal.
\newblock Lxmert: Learning cross-modality encoder representations from
  transformers.
\newblock {\em arXiv preprint arXiv:1908.07490}, 2019.

\bibitem{vaswani2017attention}
Ashish Vaswani, Noam Shazeer, Niki Parmar, Jakob Uszkoreit, Llion Jones,
  Aidan~N Gomez, {\L}ukasz Kaiser, and Illia Polosukhin.
\newblock Attention is all you need.
\newblock In {\em Advances in neural information processing systems}, pages
  5998--6008, 2017.

\bibitem{wang2022object}
Jinpeng Wang, Yixiao Ge, Guanyu Cai, Rui Yan, Xudong Lin, Ying Shan, Xiaohu
  Qie, and Mike~Zheng Shou.
\newblock Object-aware video-language pre-training for retrieval.
\newblock In {\em Proceedings of the IEEE/CVF Conference on Computer Vision and
  Pattern Recognition}, pages 3313--3322, 2022.

\bibitem{wang2020minilm}
Wenhui Wang, Furu Wei, Li Dong, Hangbo Bao, Nan Yang, and Ming Zhou.
\newblock Minilm: Deep self-attention distillation for task-agnostic
  compression of pre-trained transformers.
\newblock {\em Advances in Neural Information Processing Systems},
  33:5776--5788, 2020.

\bibitem{wang2019vatex}
Xin Wang, Jiawei Wu, Junkun Chen, Lei Li, Yuan-Fang Wang, and William~Yang
  Wang.
\newblock Vatex: A large-scale, high-quality multilingual dataset for
  video-and-language research.
\newblock In {\em Proceedings of the IEEE/CVF Conference on Computer Vision and
  Pattern Recognition}, 2019.

\bibitem{xu2017video}
Dejing Xu, Zhou Zhao, Jun Xiao, Fei Wu, Hanwang Zhang, Xiangnan He, and Yueting
  Zhuang.
\newblock Video question answering via gradually refined attention over
  appearance and motion.
\newblock In {\em Proceedings of the 25th ACM international conference on
  Multimedia}, pages 1645--1653, 2017.

\bibitem{xu2021videoclip}
Hu Xu, Gargi Ghosh, Po-Yao Huang, Dmytro Okhonko, Armen Aghajanyan, Florian
  Metze, Luke Zettlemoyer, and Christoph Feichtenhofer.
\newblock Videoclip: Contrastive pre-training for zero-shot video-text
  understanding.
\newblock {\em arXiv preprint arXiv:2109.14084}, 2021.

\bibitem{yang2021just}
Antoine Yang, Antoine Miech, Josef Sivic, Ivan Laptev, and Cordelia Schmid.
\newblock Just ask: Learning to answer questions from millions of narrated
  videos.
\newblock In {\em Proceedings of the IEEE/CVF International Conference on
  Computer Vision}, pages 1686--1697, 2021.

\bibitem{yang2022learning}
Antoine Yang, Antoine Miech, Josef Sivic, Ivan Laptev, and Cordelia Schmid.
\newblock Learning to answer visual questions from web videos.
\newblock {\em IEEE Transactions on Pattern Analysis \& Machine Intelligence},
  (01):1--1, 2022.

\bibitem{yang2022zero}
Antoine Yang, Antoine Miech, Josef Sivic, Ivan Laptev, and Cordelia Schmid.
\newblock Zero-shot video question answering via frozen bidirectional language
  models.
\newblock {\em arXiv preprint arXiv:2206.08155}, 2022.

\bibitem{yang2020bert}
Zekun Yang, Noa Garcia, Chenhui Chu, Mayu Otani, Yuta Nakashima, and Haruo
  Takemura.
\newblock Bert representations for video question answering.
\newblock In {\em The IEEE Winter Conference on Applications of Computer
  Vision}, pages 1556--1565, 2020.

\bibitem{yu2021learning}
Weijiang Yu, Haoteng Zheng, Mengfei Li, Lei Ji, Lijun Wu, Nong Xiao, and Nan
  Duan.
\newblock Learning from inside: Self-driven siamese sampling and reasoning for
  video question answering.
\newblock In M. Ranzato, A. Beygelzimer, Y. Dauphin, P.S. Liang, and J.~Wortman
  Vaughan, editors, {\em Advances in Neural Information Processing Systems},
  volume~34, pages 26462--26474. Curran Associates, Inc., 2021.

\bibitem{yu2019activitynetqa}
Zhou Yu, Dejing Xu, Jun Yu, Ting Yu, Zhou Zhao, Yueting Zhuang, and Dacheng
  Tao.
\newblock Activitynet-qa: A dataset for understanding complex web videos via
  question answering.
\newblock In {\em Proceedings of the IEEE/CVF Conference on Computer Vision and
  Pattern Recognition}, pages 9127--9134, 2019.

\bibitem{zellers2021merlot}
Rowan Zellers, Ximing Lu, Jack Hessel, Youngjae Yu, Jae~Sung Park, Jize Cao,
  Ali Farhadi, and Yejin Choi.
\newblock Merlot: Multimodal neural script knowledge models.
\newblock {\em Advances in Neural Information Processing Systems}, 34, 2021.

\bibitem{zhou2020unified}
Luowei Zhou, Hamid Palangi, Lei Zhang, Houdong Hu, Jason~J Corso, and Jianfeng
  Gao.
\newblock Unified vision-language pre-training for image captioning and vqa.
\newblock {\em arXiv preprint arXiv:1909.11059}, 2019.

\bibitem{youcook2}
Luowei Zhou, Chenliang Xu, and Jason~J Corso.
\newblock Towards automatic learning of procedures from web instructional
  videos.
\newblock In {\em Thirty-Second AAAI Conference on Artificial Intelligence},
  2018.

\bibitem{ZhXuCoAAAI18}
Luowei Zhou, Chenliang Xu, and Jason~J Corso.
\newblock Towards automatic learning of procedures from web instructional
  videos.
\newblock In {\em AAAI Conference on Artificial Intelligence}, pages
  7590--7598, 2018.

\end{thebibliography}
}

\appendix
\section{Discussion on Reasons Why Text + Text Performs Best}
We think the main reason that the proposed Text + Text variant can outperform other variants is that it is hard and resource-intensive to learn the alignment between the text embedding space and the visual feature space, and train a good multimodal transformer even with strong pretrained text models as initialization. There are two lines of evidence we observed: first, as discussed in the next section, these two spaces are significantly different even in terms of statistics, which is verified by the fact that without the trick to initialize the projected visual features to have the same mean and variance with text embeddings, Conti. + Text would suffer from a huge performance drop of 4\%; second, from results in the main paper, it is clear that Conti. + Multi. seriously rely on the huge amount of video-text samples to align the embedding space and train the multimodal transformer. The Text + Text variant exactly avoids these two processes in its design.

\section{Additional Implementation Details}
We use a learning rate of 0.00005, learning rate decay of 0.9 and batch size of 256. We trained our model for 20 epochs on iVQA and MSVD-QA and for 30 epochs on How2QA, ActivityNet-QA and MSRVTT-QA on 2 Nvidia V100 GPUs. On VATEX and YouCook II, the number epochs is 10 and 100, respectively. We find that the Masked Language Modeling objective is not helpful when we use MPNet as the language transformer, and thus, we only use the contrastive loss as described in the main paper. We use a gradient clipping of 1, following~\cite{n2019sbert}. The other hyper-parameters were directly borrowed from Just-ask~\cite{yang2021just}.

In the implementation of Continuous Features + Text Transformer, we find it important to initialize the last LayerNorm~\cite{ba2016layer} in the projector with parameters of the LayerNorm after the embedding layer of MPNet so that the projected video features are aligned with the textual embeddings. Without this initialization trick, the model only achieves 19.8\% on iVQA (when with the initialization, it achieves 23.2\%).

We use a public python library\footnote{\url{https://pypi.org/project/youtube-transcript-api/}} to obtain the automatic speech transcripts from YouTube for the videos we used in iVQA, How2QA, ActivityNet-QA, YouCook II and VATEX. We also used subtitles from~\cite{li2021value} for videos with speech but without ASR. Eventually, about 90\%, 70\%, 30\%, 90\%, and 50\% of the videos have associated speech transcripts, respectively.

We fix the random seed in all the experiments and we do not observe significant change of accuracy ($<0.5\%$) when changing the random seed. The code will be released at \url{https://github.com/XudongLinthu/upgradable-multimodal-intelligence} for any other details and results on additional datasets.

\section{Details about Datasets and Evaluation Metric}
\noindent \textbf{iVQA~\cite{yang2021just}}. It contains 10,000 instructional videos. Each video is annotated with one question and five corresponding answers. We follow the official split to use 6,000, 2,000, and 2,000 videos for training, validation, and testing, respectively. We follow~\cite{yang2021just} to calculate accuracy with five annotations per question. 

\noindent \textbf{How2QA~\cite{li2020hero}}. The dataset contains 44,007 QA pairs that are annotated from 9,035 videos. We follow~\cite{yang2021just} to use the train and validation split for training and testing. Note that in this dataset, each question and answer pair are manually annotated with three negative answers so we actually don't need to retrieve from a huge answer set. The metric used for this dataset is accuracy.

\noindent \textbf{ActivityNet-QA~\cite{yu2019activitynetqa}}. It contains 58,000 QA pairs manually annotated from 5,800 videos from the ActivityNet \cite{s2016activitynet} dataset depicting a wide range of complex human activities. The official split of 32,000, 18,000 and 8,000 QA pairs for training, validation and testing respectively is adopted. 

\noindent \textbf{YouCook II~\cite{ZhXuCoAAAI18}}. It is a instructional video dataset containing 2,000 long videos of 89 recipes. We follow~\cite{miech2020end} to use the temporal boundary of steps to formulate a pair as a step description and the corresponding video segment. The resulted number of training and testing pairs are 10,387 and 3,411.

\noindent \textbf{VATEX~\cite{wang2019vatex}}. We only take videos and English captions from it to evaluate retrieval performance. Due to the fact that only 50\% of the videos have ASR and many ASRs only contain English stop words, we only keep the videos with at least 5 non-stop words to make sure the task is still multi-channel. The resulted number of training and testing pairs are 36,680 and 4,190. When comparing with HERO~\cite{li2020hero}, we train and evaluate our model under its data split.




\begin{table}[t]

\centering

\small
\begin{tabu}{ccc}

\toprule
Number of Tokens &  60k-word   &  Answer-word

  
\\ \midrule  
10 &   25.9  &  29.8 

\\ \midrule  
15 &   27.3   &  30.9

\\ \midrule  
20 &   26.8  &  31.1

\\ \midrule  
25 &  26.5    &  31.6

\\ \midrule  
30 &  26.2  &  30.9  \\

\bottomrule
\end{tabu}
\caption{Accuracy on varying the number of text tokens and 60k-word vocabulary and the answer-word vocabulary on the iVQA dataset }

\label{tokens}

\end{table}

\section{Experiments on \textit{k} and the Vocabulary.}
In this section, we report the results when varying $k$ and the vocabulary for retrieving text tokens, as shown in Table~\ref{tokens}. 60k-word vocabulary contains all 65,000 the words that are used in the language model of~\cite{miech2020end}. The answer-word vocabulary is constructed by collecting all the unique verbs and nouns from parsing all the query/answer sentences in the downstream datasets with spaCy~\cite{honnibal2017spacy}.

We observe that when larger than 15 words are retrieved for each segment in the video, the benefit of retrieving more tokens starts to be marginal. Therefore, we use $k=15$ for all the other datasets as less tokens also help to further accelerate the training process. But we use $k=25$ for the iVQA dataset as this turns out to be the optimal value when using the answer-word vocabulary on the iVQA dataset. This also indicates that the performance on other downstream datasets could be further improved if we optimize the number of tokens. 

We also observe a consistent improvement for all the number of tokens when changing from the 60k-word vocabulary to the answer-word vocabulary. Therefore, we construct the answer-word vocabulary for each dataset separately, which have a size from 3K to 25K words, depending on the dataset. Note that for retrieval datasets, we use the words from all the text queries.

\begin{table*}[t]

\centering

\small
\begin{tabu}{ccccc}

\toprule
Model &  Extra MM Samples   & $\Delta$ GPU hours & YouCook II  & VATEX
\\ \midrule
\rowfont{\color{gray}} AT-ST HERO~\cite{li2021value,li2020hero} & 7.6M + 700K & - & 45.3 & 80.0 \\
\midrule
 HERO~\cite{li2021value,li2020hero} & 7.6M        & ~8,000 & 49.5  & \textbf{63.4} \\
  
\midrule  
 Text + Text (Ours)  & 0 & 0   &   \textbf{50.6}  & 60.1 \\

\bottomrule
\end{tabu}
\caption{Comparison with the state-of-the-art on YouCook II and VATEX in terms of accuracy and efficiency. Extra MM Samples indicate the number of video-text samples that are needed in the second-round pretraining. $\Delta$ GPU hours refer to the additional computation required for the second-round pretraining. Note that our variant typically requires 1 GPU hour for training. Average recall@\{1,5,10\} (\%) is reported. AT-ST HERO is annotated as gray because it further leverages about 730K extra \textbf{annotated} multimodal samples to perform multi-task training, which is not directly comparable to our setting. Note that even HERO is \text{not completely comparable} with our model, as it enjoys both video-level and frame-level feature extractors.
}

\label{sota}
\end{table*}

\section{Additional Comparison with the State-of-the-art}
As shown in Table~\ref{sota}, we provide additional comparison with state-of-the-art on the two retrieval datasets. We observe that our proposed \textbf{Text Tokens + Text Transformer} performs slightly better than HERO, which requires a lot more data and computational resources to train the model properly. Note that our model only uses pretrained S3D~\cite{miech2020end} (on YouCook II) or CLIP~\cite{radford2021learning} (on VATEX)  to retrieve text tokens as additional representation of the video but HERO that uses features from both 3D extractors~\cite{miech2020end,feichtenhofer2019slowfast} and 2D extractors~\cite{radford2021learning}. Note that we found that on YouCook II, using $K=25$ helps to improve the results. 

AT-ST HERO is annotated as gray because it further leverages about 730K extra \textbf{annotated} multimodal samples to perform multi-task training and then it is fine-tuned for specific datasets, which is not directly comparable to our setting. Interestingly, the additional annotated data helps AT-ST HERO to significantly improves the performance on VATEX but the performance on YouCook II is actually even lower. Overall, when without such multi-task learning, our proposed method still achieves comparable results with state-of-the-art on these datasets.

\section{Additional Discussion on MSRVTT-QA and MSVD-QA}
We list a few question-answer samples from the test set of MSRVTT-QA in Table~\ref{msrvtt}. The questions are not natural sentences and the answers are not informative.
To quantify the imperfection of question generation, we carefully convey a manual study on 50 randomly sampled question-answer-video triplets in the MSRVTT-QA dataset. We found that the 6\% of the questions are not exactly aligned with the video, e.g., wrong entity descriptions/wrong action descriptions. 24\% of the questions have grammatical errors. 10\% of the questions are ambiguous. 
For the answers, we found that 12\% of them are either not aligned with the video or too ambiguous to determine correctness. 32\% of the answers are not informative enough but they are roughly aligned with the question and the video.

A side evidence is that the Table 7 in \cite{yang2021just} compares between the method used to generate MSRVTT-QA and MSVD-QA and the method used in Just-ask for question-answer generation. The resulted model pretrained with generation method of \cite{yang2021just} significantly outperforms the model pretrained with data generated from method \cite{heilman2010good}, by a large relative margin of 20\% to 1000\% of zero-shot accuracy on three manually annotated datasets, which indicates the unsatisfactory quality of MSRVTT-QA and MSVD-QA generated by \cite{heilman2010good}.

Both the qualitative and quantitative results motivate us to not use them as the datasets to assess the performance of the four variants.

\begin{table}[th]

\centering

\small
\begin{tabu}{cc}

\toprule
Question & Answer
\\ \midrule
\midrule
what is a clip doing? & show \\
who is showing something in a computer? & someone \\
what can singing? &   pain \\
what explains colors? & video \\
\bottomrule
\end{tabu}

\caption{Question-answer samples from the test set of MSRVTT-QA.
}
\label{msrvtt}
\end{table}

\section{Additional Visualisation}
\begin{figure*}[h]
    \centering
    \includegraphics[width=0.99\linewidth]{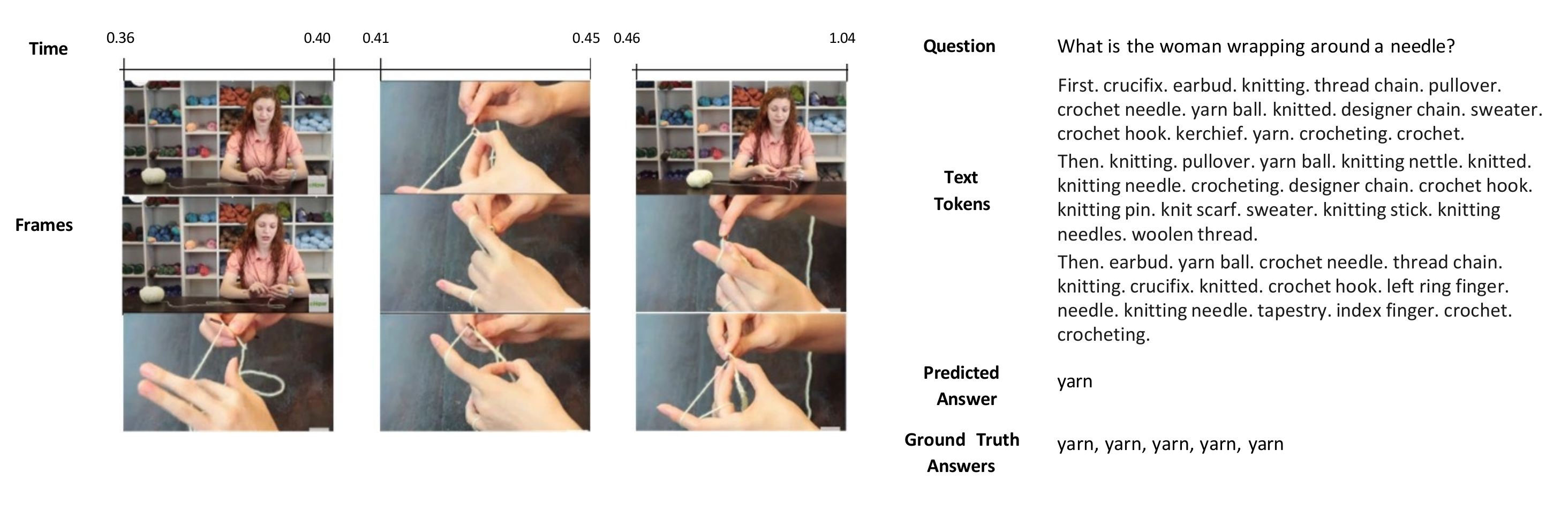}
    \caption{Visualisation of a successful case for \textbf{Text Tokens + Text Transformer} on the iVQA dataset. In this example, the answer ``yarn'' is retrieved as one of the text tokens to describe the video.}
    \label{fig:vis3}
\end{figure*}

As shown in Figure~\ref{fig:vis3}, the answer words or some of the answer words are retrieved as one of the text tokens to describe the video. As discussed in the main paper, we find that 64\% of the videos have at least one word overlap between the retrieved text tokens and the answers. This kind of examples show the high explainability of the proposed method as it is clear these important key words are the input to the model to make the final prediction.

\section{Discussion on Social Impact, Limitation and Future Work}
We first argue that the tasks we handle in this work are potentially helpful for visually-impaired people to better handle daily life as our model can be used to help them understand the ongoing events with text queries.

The datasets we used are mostly based on videos from YouTube. Therefore, they may contain personal information but our algorithm is not designed to specifically leverage certain private information. Overall, we do not expect negative societal impact from the designed algorithms but the dataset we use for training may lead the models to produce biased or undesired results.

One limitation is on the imperfection of the visual words retrieval process. We have already shown using CLIP~\cite{radford2021learning} significantly helps to improve the performance in the main paper. We leave further using better pretrained multimodal contrastive models to future work. We also leave the utilization of recent larger pretrained contrastive text models~\cite{ni2021large} (with billions of parameters) to future research.

\end{document}